\documentclass[11pt,a4paper]{article}
\usepackage[hyperref]{naaclhlt2018}
\usepackage{times}
\usepackage{latexsym}
\usepackage{graphicx}
\usepackage{amsmath}
\usepackage{booktabs}
\usepackage{url}
\usepackage{bm}
\usepackage[multiple]{footmisc}

\aclfinalcopy 

\title{Training a Ranking Function for Open-Domain Question Answering} 

\author{Phu Mon Htut$^1$ \\
  {\tt pmh330@nyu.edu} \\\And
  Samuel R. Bowman$^{1,2,3}$ \\
  {\tt bowman@nyu.edu} \\\And
  Kyunghyun Cho$^{1,3}$ \\
  {\tt kyunghyun.cho@nyu.edu} \\
  \AND
$^{1}$\normalfont Center for Data Science\\New York University\\60 Fifth Avenue\\New York, NY 10011\And
$^{2}$\normalfont Dept. of Linguistics\\New York University\\10 Washington Place\\New York, NY 10003\And
$^{3}$\normalfont Dept. of Computer Science\\New York University\\60 Fifth Avenue\\New York, NY 10011
  } 
  
\date{}

\begin{document}
\maketitle
\begin{abstract}
  In recent years, there have been amazing advances in deep learning methods for machine reading. In machine reading, the machine reader has to extract the answer from the given ground truth paragraph. Recently, the state-of-the-art machine reading models achieve human level performance in SQuAD which is a reading comprehension-style question answering (QA) task.
The success of machine reading has inspired researchers to combine information retrieval with machine reading to tackle open-domain QA.
However, these systems perform poorly compared to reading comprehension-style QA because it is difficult to retrieve the pieces of paragraphs that contain the answer to the question. In this study, we propose two neural network rankers that assign scores to different passages based on their likelihood of containing the answer to a given question. Additionally, we analyze the relative importance of semantic similarity and word level relevance matching in open-domain QA.
\end{abstract}

\begin{figure*}[t!]
\center
\includegraphics[width=0.8\textwidth]{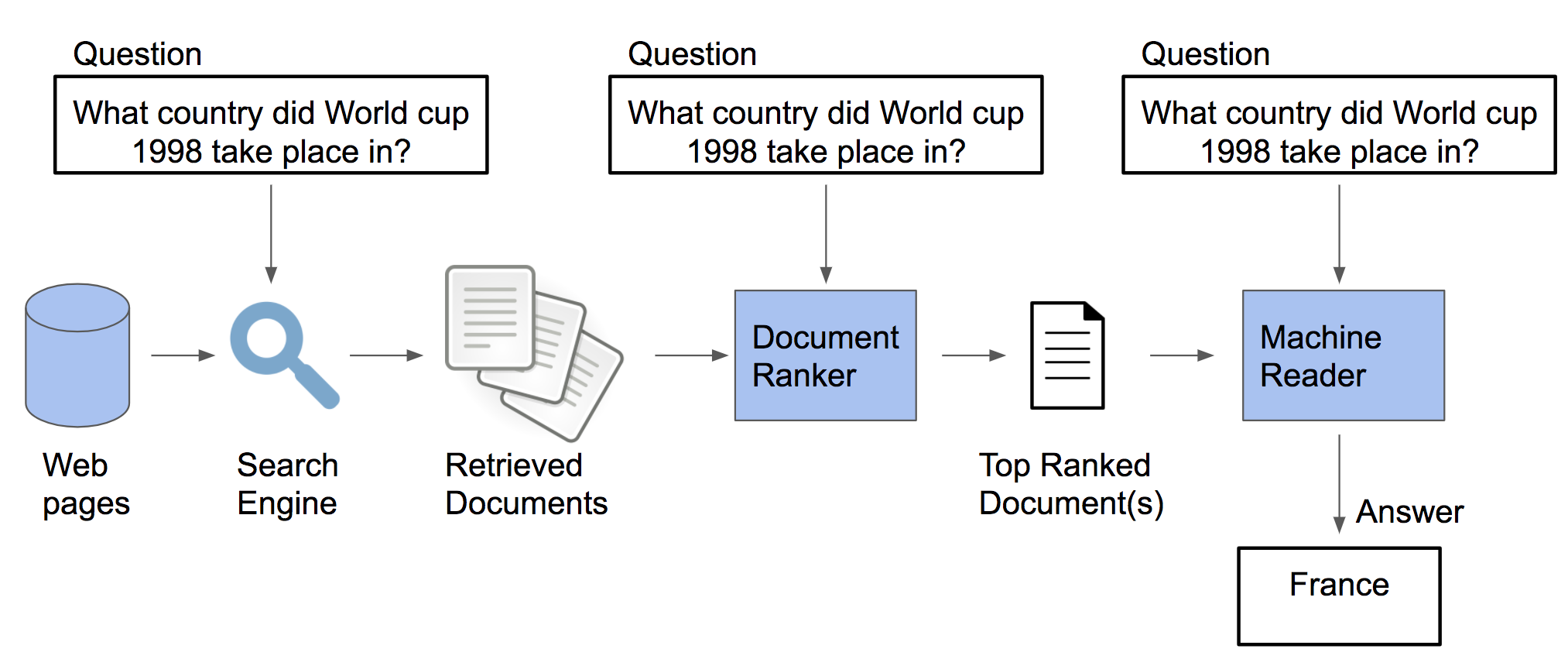} 
\caption{The overall pipeline of the open-domain QA model}
\label{architecture}
\end{figure*} 

\section{Introduction}  
The goal of a question answering (QA) system is to provide a relevant answer to a natural language question. In reading comprehension-style QA, the ground truth paragraph that contains the answer is given to the system whereas no such information is available in open-domain QA setting. Open-domain QA systems have generally been built upon large-scale structured knowledge bases, such as Freebase or DBpedia. The drawback of this approach is that these knowledge bases are not complete~\cite{DBLP:conf/www/WestGMSGL14}, and are expensive to construct and maintain.

Another method for open-domain QA is a corpus-based approach where the QA system looks for the answer in the unstructured text corpus~\cite{DBLP:conf/trec/BrillLBDN01}. This approach eliminates the need to build and update knowledge bases by taking advantage of the large amount of text data available on the web. Complex parsing rules and information extraction methods are required to extract answers from unstructured text. As machine readers are excellent at this task, there have been attempts to combine search engines with machine reading for corpus-based open-domain QA \citep{DBLP:conf/acl/ChenFWB17,DBLP:journals/corr/abs-1709-00023}. To achieve high accuracy in this setting, the top documents retrieved by the search engine must be relevant to the question. As the top ranked documents returned from search engine might not contain the answer that the machine reader is looking for, re-ranking the documents based on the likelihood of containing answer will improve the overall QA performance. Our focus is on building a neural network ranker to re-rank the documents retrieved by a search engine to improve overall QA performance. 

Semantic similarity is crucial in QA as the passage containing the answer may be semantically similar to the question but may not contain the exact same words in the question. For example, the answer to ``\textit{What country did world cup 1998 take place in?}'' can be found in ``\textit{World cup 1998 was held in France}.'' Therefore, we evaluate the performance of the fixed size distributed representations that encode the general meaning of the whole sentence on ranking. We use a simple feed-forward neural network with fixed size question and paragraph representations for this purpose.

In ad-hoc retrieval, the system aims to return a list of documents that satisfies the user's information need described in the query.\footnote{Information Retrieval Glossary: http://people.ischool.berkeley.edu/~hearst/irbook/glossary.html} \citet{DBLP:journals/corr/abs-1711-08611} show that, in ad-hoc retrieval, relevance matching---identifying whether a document is relevant to a given query---matters more than semantic similarity. Unlike semantic similarity, that measures the overall similarity in meaning between a question and a document, relevance matching measures the word or phrase level local interactions between pieces of texts in a question and a document. As fixed size representations encode the general meaning of the whole sentence or document, they lose some distinctions about the keywords that are crucial for retrieval and question answering. To analyze the importance of relevance matching in QA, we build another ranker model that focuses on local interactions between words in the question and words in the document.
We evaluate and analyze the performance of the two rankers on QUASAR-T dataset~\cite{dhingra2017quasar}. We observe that the ranker model that focuses on relevance matching (Relation-Networks ranker) achieves significantly higher retrieval recall but the ranker model that focuses on semantic similarity (InferSent ranker) has better overall QA performance. We achieve 11.6 percent improvement in overall QA performance by integrating InferSent ranker (6.4 percent improvement by Relation-Networks ranker).

\section{Related Work}
With the introduction of large-scale datasets for machine reading such as CNN/DailyMail~\cite{DBLP:conf/nips/HermannKGEKSB15} and The Stanford Question Answering Dataset \citep[SQuAD;][]{DBLP:conf/emnlp/RajpurkarZLL16},  the machine readers have become increasingly accurate at extracting the answer from a given paragraph. 
In machine reading-style question answering datasets like SQuAD, the system has to locate the answer to a question in the given ground truth paragraph. Neural network based models excel at this task and have recently achieved human level accuracy in SQuAD.\footnote{SQuAD leaderboard: https://rajpurkar.github.io/SQuAD-explorer/}

Following the advances in machine reading, researchers have begun to apply Deep Learning in corpus-based open-domain QA approach by incorporating information retrieval and machine reading. \citet{DBLP:conf/acl/ChenFWB17} propose a QA pipeline named DrQA that consists of a Document Retriever and a Document Reader. The Document Retriever is a TF-IDF retrieval system built upon Wikipedia corpus. The Document Reader is a neural network machine reader trained on SQuAD. Although DrQA's Document Reader achieves the exact match accuracy of 69.5 in reading comprehension-style QA setting of SQuAD, their accuracy drops to 27.1 in the open-domain setting, when the paragraph containing the answer is not given to the reader. In order to extract the correct answer, the system should have an effective retrieval system that can retrieve highly relevant paragraphs. Therefore, retrieval plays an important role in open-domain QA and current systems are not good at it. 

To improve the performance of the overall pipeline, \citet{DBLP:journals/corr/abs-1709-00023} propose Reinforced Ranker-Reader ($R^3$) model. The pipeline of $R^3$ includes a Lucene-based search engine, and a neural network ranker that re-ranks the documents retrieved by the search engine, followed by a machine reader. The ranker and reader are trained jointly using reinforcement learning. Qualitative analysis of top ranked documents by the ranker of $R^3$ shows that the neural network ranker can learn to rank the documents based on semantic similarity with the question as well as the likelihood of extracting the correct answer by the reader. 

We follow a similar pipeline as \citet{DBLP:journals/corr/abs-1709-00023}. Our system consists of a neural network ranker and a machine reader as shown in Figure 1. The focus of our work is to improve the ranker for QA performance. We use DrQA's Document Reader as our reader. We train our ranker and reader models on QUASAR-T \citep{dhingra2017quasar} dataset. QUASAR-T provides a collection top 100 short paragraphs returned by search engine for each question in the dataset. Our goal is to find the correct answer span for a given question.

\section{Model Architecture}

\subsection{Overall Setup}
The overall pipeline consists of a search engine, ranker and reader. We do not build our own search engine as QUASAR-T provides 100 short passages already retrieved by the search engine for each question. We build two different rankers: \textbf{InferSent ranker} to evaluate the performance of semantic similarity in ranking for QA, and \textbf{Relation-Networks ranker} to evaluate the performance of relevance matching in ranking for QA. We use the Document Reader of DrQA~\cite{DBLP:conf/acl/ChenFWB17} as our machine reader.

\subsection{Ranker}
Given a question and a paragraph, the ranker model acts as a scoring function that calculates the similarity between them. In our experiment, we explore two neural network models as the scoring functions of our rankers: a feed-forward neural network that uses InferSent sentence representations~\cite{DBLP:conf/emnlp/ConneauKSBB17} and Relation-Networks~\cite{DBLP:conf/nips/SantoroRBMPBL17}. We train the rankers by minimizing the margin ranking loss~\cite{DBLP:journals/ir/BaiWGCSQCW10}: 
\begin{equation}
\sum_{i=1}^{k} max(0, 1 - f(q,p_{pos}) + f(q,p_{neg}^{i})) 
\end{equation}
where $f$ is the scoring function, $p_{pos}$ is a paragraph that contains the ground truth answer, $p_{neg}$ is a negative paragraph that does not contain the ground truth answer, and $k$ is the number of negative paragraphs. We declare paragraphs that contain the exact ground truth answer string provided in the dataset as positive paragraphs. For every question, we sample one positive paragraph and five negative paragraphs. Given a question and a list of paragraphs, the ranker will return the similarity scores between the question and each of the paragraphs.

\subsubsection{InferSent Ranker}

InferSent~\cite{DBLP:conf/emnlp/ConneauKSBB17} provides distributed representations for sentences.\footnote{https://github.com/facebookresearch/InferSent} It is trained on Stanford Natural Language Inference Dataset \citep[SNLI; ][]{snli:emnlp2015} and Multi-Genre NLI Corpus \citep[MultiNLI; ][]{DBLP:journals/corr/WilliamsNB17} using supervised learning. It generalizes well and outperforms unsupervised sentence representations such as Skip-Thought Vectors~\cite{DBLP:conf/nips/KirosZSZUTF15} in a variety of tasks. 

As InferSent representation captures the general semantics of a sentence, we use it to implement the ranker that ranks based on semantic similarity. To compose sentence representations into a paragraph representation, we simply sum the InferSent representations of all the sentences in the paragraph. This approach is inspired by the sum of word representations as composition function for forming sentence representations \citep{DBLP:conf/acl/IyyerMBD15}.

We implement a feed-forward neural network as our scoring function. The input feature vector is constructed by concatenating the question embedding, paragraph embedding, their difference, and their element-wise product \citep{Mou:16}:

\begin{align}
    \bm{x}_{classifier} &= \begin{bmatrix}
           \bm{q} \\
           \bm{p} \\
           \bm{q} - \bm{p} \\
           \bm{q} \bigodot \bm{p}
         \end{bmatrix}\\
   \bm{z} &= \bm{W}^{(1)}\bm{x}_{classifier} + \bm{b}^{(1)}\\
   score &= \bm{W}^{(2)} ReLU(\bm{z}) + \bm{b}^{(2)}
\end{align}
The neural network consists of a linear layer followed by a ReLU activation function, and another scalar-valued linear layer that provides the similarity score between a question and a paragraph.

\subsubsection{Relation-Networks (RN) Ranker}
We use Relation-Networks ~\cite{DBLP:conf/nips/SantoroRBMPBL17} as the ranker model that focuses on measuring the relevance between words in the question and words in the paragraph. Relation-Networks are designed to infer the relation between object pairs. In our model, the object pairs are the question word and context word pairs as we want to capture the local interactions between words in the question and words in the paragraph. The word pairs will be used as input to the Relation-Networks:
\begin{equation}
RN(q,p) = f_{\phi}\bigg(\sum_{i,j} g_{\theta}([E(q_i);E(p_j)]) \bigg) 
\end{equation}
where $q = \{q_1, q_2,..., q_n\}$ is the question that contains n words and $p = \{p_1, p_2,..., p_m\}$ is the paragraph that contains m words; $E(q_i)$ is a 300 dimensional GloVe embedding~\cite{DBLP:conf/emnlp/PenningtonSM14} of word $q_i$, and $[\cdot;\cdot]$ is the concatenation operator. $f_{\phi}$ and $g_{\theta}$ are 3 layer feed-forward neural networks with ReLU activation function. 

The role of $g_{\theta}$ is to infer the relation between two words while $f_{\phi}$ serves as the scoring function. As we directly compare the word embeddings, this model will lose the contextual information and word order, which can provide us some semantic information. We do not fine-tune the word embeddings during training as we want to preserve the generalized meaning of GloVe embeddings. We hypothesize that this ranker will achieve a high retrieval recall as relevance matching is important for information retrieval \citep{DBLP:journals/corr/abs-1711-08611}. 

\begin{table*}[t!]
\centering
\begin{tabular}{*{2}{p{.47\linewidth}}}
\toprule
\multicolumn{2}{c}{ Question: Which country's name means ``equator"? Answer: Ecuador}\\ \midrule
 \multicolumn{1}{c}{\textbf{InferSent ranker}} & \multicolumn{1}{c}{\textbf{RN ranker}}\\ \midrule
  \textbf{Ecuador} : ``Equator" in Spanish , as the country lies on the Equator. & Salinas, is considered the best tourist beach resort in Ecuador's Pacific Coastline.. Quito, \textbf{Ecuador} Ecuador's capital and the country's second largest city.  \\ 
  The equator crosses just north of \textbf{Ecuador}'s capital, Quito, and the country gets its name from this hemispheric crossroads. & Quito is the capital of \textbf{Ecuador} and of Pichincha, the country's most populous Andean province, is situated 116 miles from the Pacific coast at an altitude of 9,350 feet, just south of the equator.  \\
  The country that comes closest to the equator without actually touching it is Peru. & The location of the Republic of \textbf{Ecuador} Ecuador, known officially as the Republic of Ecuador -LRB- which literally means ``Republic of the equator" -RRB- , is a representative democratic republic \\
  The name of the country is derived from its position on the Equator. & The name of the country is derived from its position on the Equator. \\ 
\bottomrule
\end{tabular}
\caption{An example question from the QUASAR-T test set with the top passages returned by the two rankers. 
  } 
\end{table*}

\subsection{Machine Reader}
The Document Reader of DrQA \citep{DBLP:conf/acl/ChenFWB17} is a multi-layer recurrent neural network model that is designed to extract an answer span to a question from a given document (or paragraphs). We refer readers to the original work for details. We apply the default configuration used in the original work, and train the DrQA on QUASAR-T dataset.\footnote{DrQA code available at: https://github.com/facebookresearch/DrQA .} 

\subsection{Paragraph Selection}
The ranker provides the similarity score between a question and each paragraph in the article. We select the top 5 paragraphs based on the scores provided by the ranker. We use soft-max over the top 5 scores to find $P(\text{$p^i_j$})$, the model's estimate of the probability that the passage is the most relevant one from among the top 5.

Furthermore, the machine reader provides the probability of each answer span given a paragraph $P(\text{$answer_j$}|\text{$p^i_j$})$, where $answer_j$ stands for the answer span of $j^{th}$ question in dataset and $p^i_j$ indicates the corresponding top 5 paragraphs. We can thus calculate the overall confidence of each answer span and corresponding paragraph $P(\text{$p^i_j$},\text{$answer_j$})$ by multiplying $P(\text{$answer_j$}|\text{$p^i_j$})$ with $P(\text{$p^i_j$})$.
We then choose the answer span with the highest $P$($answer_j$, $p^i_j$) as the output of our model.

\section{Experimental Setup}

\subsection{QUASAR Dataset}
The QUestion Answering by Search And Reading (QUASAR) dataset~\cite{dhingra2017quasar} includes QUASAR-S and QUASAR-T, each designed to address the combination of retrieval and machine reading. QUASAR-S consists of fill-in-the-gaps questions collected from Stackoverflow using software entity tags. As our model is not designed for fill-in-the-gaps questions, we do not use QUASAR-S. QUASAR-T, which we use, consists of 43,013 open-domain questions based on trivia, collected from various internet sources. The candidate passages in this dataset are collected from a Lucene based search engine built upon ClueWeb09.\footnote{https://lucene.apache.org/}\footnote{https://lemurproject.org/clueweb09/}

\subsection{Baselines}
We consider four models with publicly available results for QUASAR-T dataset. GA: Gated Attention Reader~\cite{DBLP:conf/acl/DhingraLYCS17}, BiDAF: Bidirectional Attention Flow~\cite{DBLP:journals/corr/SeoKFH16}, $R^3$: Reinforced Ranker-Reader~\cite{DBLP:journals/corr/abs-1709-00023} and $SR^2$: Simple Ranker-Reader~\cite{DBLP:journals/corr/abs-1709-00023} which is a variant of $R^3$ that jointly trains the ranker and reader using supervised learning.

\subsection{Implementation Details}
Each InferSent embedding has 4096 dimensions. Therefore, the input feature vector to our InferSent ranker has 16384 dimensions. The dimensions of the two linear layers are 500 and 1.

As for Relation-Networks (RN), $g_{\theta}$ and $f_{\phi}$ are three layer feed-forward neural networks with (300, 300, 5) and (5, 5, 1) units respectively. 

We use NLTK to tokenize words for the RN ranker.~\footnote{https://www.nltk.org/} We lower-case the words, and remove punctuations and infrequent words that occur less than 5 times in the corpus. We pass untokenized sentence string as input directly to InferSent encoder as expected by it.

We train both the InferSent ranker and the RN ranker using Stochastic Gradient Descent with the Adam optimizer~\cite{DBLP:journals/corr/KingmaB14}.\footnote{Learning rate is set to 0.001.}  A dropout \citep{DBLP:journals/jmlr/SrivastavaHKSS14} of p=0.5 is applied to all hidden layers for training both InferSent and RN rankers.

\section{Results and Analysis}

First, we evaluate the two ranker models based on the recall@K, which measures whether the ground truth answer span is in the top $K$ ranked documents. We then evaluate the performance of machine reader on the top $K$ ranked documents of each ranker by feeding them to DrQA reader and measuring the exact match accuracy and F-1 score produced by the reader. Finally, we do qualitative analysis of top-5 documents produced by each ranker.

\subsection{Recall of Rankers}
The performance of the rankers is shown in Table 2. 
Although the recall of the InferSent ranker is somewhat lower than that of the $R^3$ ranker at top-1, it still improves upon the recall of search engine provided by raw dataset. In addition, it performs slightly better than the ranker from $R^3$ for recall at top-3 and top-5. We can conclude that using InferSent as paragraph representation improves the recall in re-ranking the paragraphs for machine reading. 

The RN ranker achieves significantly higher recall than $R^3$ and InferSent rankers. This proves our hypothesis that word by word relevance matching improves retrieval recall. Does high recall mean high question answering accuracy? We further analyze the documents retrieved by the rankers to answer this.

\subsection{Machine Reading Performance}
For each question, we feed the top five documents retrieved by each ranker to DrQA trained on QUASAR-T to produce an answer. The overall QA performance improves from exact match accuracy of 19.7 to 31.2 when InferSent ranker is used (Table 3). We can also observe that InferSent ranker is much better than RN ranker in terms of overall QA performance despite its low recall for retrieval. InferSent ranker with DrQA provides comparable result to $SR^2$ despite being a simpler model.

\begin{table}
\begin{center}
\begin{tabular}{lccc}
\toprule
 & \bf Top-1 & \bf Top-3 & \bf Top-5 \\ \midrule
IR & 19.7 & 36.3 & 44.3 \\
Ranker from $R^3$ & 40.3 & 51.3 & 54.5 \\ \midrule
InferSent ranker & 36.1 & 52.8 & 56.7 \\
RN ranker & \bf 51.4 & \bf 68.2 & \bf 70.3 \\
\bottomrule 
\end{tabular}
\end{center}
\caption{\label{font-table} Recall of ranker on QUASAR-T test dataset. The recall is calculated by checking whether the ground truth answer appears in top-N paragraphs. IR is the search engine ranking given in QUASAR-T dataset.}
\end{table}

\begin{table}
\begin{center}
\begin{tabular}{lcc}
\toprule
 & \bf EM & \bf F1 \\ \midrule
No ranker + DrQA & 19.6 & 24.43 \\
InferSent + DrQA & \underline{31.2} & \underline{37.6}  \\ 
RN + DrQA & 26.0 & 30.7 \\ \midrule 
GA ~\cite{DBLP:conf/acl/DhingraLYCS17} & 26.4 & 26.4 \\
BiDAF ~\cite{DBLP:journals/corr/SeoKFH16} & 25.9 & 28.5 \\
$R^3$ ~\cite{DBLP:journals/corr/abs-1709-00023} & \bf 35.3 & \bf 41.7 \\
$SR^2$ ~\cite{DBLP:journals/corr/abs-1709-00023} & 31.9 & 38.7 \\
\bottomrule
\end{tabular}
\end{center}
\caption{\label{font-table} Exact Match(EM) and F-1 scores of different models on QUASAR-T test dataset. Our InferSent + DrQA model is as competitive as {$SR^2$} which is a supervised variant of the state-of-the-art model, {$R^3$}}
\end{table}

\subsection{Analysis of paragraphs retrieved by the rankers}
The top paragraphs ranked by InferSent are generally semantically similar to the question (Table 1). However, we find that there is a significant number of cases where proper noun ground truth answer is missing in the paragraph. An example of such a sentence would be ``\textit{The name of the country is derived from its position on the Equator}''. Though this sentence is semantically similar to the question, it does not contain the proper noun answer. As InferSent encodes the general meaning of the whole sentence in a distributed representation, it is difficult for the ranker to decide whether the representation contains the important keywords for QA.

Although the top paragraphs ranked by RN ranker contain the ground truth answer, they are not semantically similar to the question. This behavior is expected as RN ranker only performs matching of words in question with words in the paragraph, and does not have information about the context and word order that is important for learning semantics. However, it is interesting to observe that RN ranker can retrieve the paragraph that contains the ground truth answer span even when the paragraph has little similarity with the question. 

In Table 1, the top paragraph retrieved by RN ranker is not semantically similar to the question. Moreover, the only word overlap between the question and paragraph is ``\textbf{country's}" which is not an important keyword. The fourth paragraph retrieved by RN ranker not only contains the word ``\textbf{country}" but also has more word overlap with the question. Despite this, RN ranker gives a lower score to the fourth paragraph as it does not contain the ground truth answer span. As RN ranker is designed to give higher ranking score to sentences or paragraphs that has the highest word overlap with the question, this behavior is not intuitive.
We notice many similar cases in test dataset which suggests that RN ranker might be learning to predict the possible answer span on its own. As RN ranker compares every word in question with every word in the paragraph, it might learn to give a high score to the word in the paragraph that often co-occurs with all the words in the question. For example, it might learn that \textit{Equador} is the word that has highest co-occurence with \textit{country}, \textit{name}, \textit{means} and \textit{equator}, and gives very high scores to the paragraphs that contain \textit{Equador}.

Nevertheless, it is difficult for machine reader to find answer in such paragraphs that have little or no meaningful overlap with the question. This explains the poor performance of machine reader on documents ranked by RN ranker despite its high recall. 

\section{Conclusion}
We find that word level relevance matching significantly improves retrieval performance. We also show that the ranker with very high retrieval recall may not achieve high overall performance in open-domain QA. Although both semantic similarity and relevance scores are important for open-domain QA, we find that semantic similarity contributes more for a better overall performance of open-domain QA. 
For the future work, we would like to explore new ranking models that consider both overall semantic similarity and weighted local interactions between words in the question and the document. Moreover, as Relation-Networks are very good at predicting the answer on their own, we would like to implement a model that can do both ranking and answer extraction based on Relation-Networks.

\section*{Acknowledgments}

PMH is funded as an AdeptMind Scholar. This project has benefited from financial support to SB by Google, Tencent Holdings, and Samsung Research. KC thanks support by AdeptMind, eBay, TenCent, NVIDIA and CIFAR. This work is partly funded by the Defense Advanced Research Projects Agency (DARPA) D3M program. Any opinions, findings, and conclusions or recommendations expressed in this material are those of the authors and do not necessarily reflect the views of DARPA. 

\bibliography{naaclhlt2018}
\bibliographystyle{acl_natbib}

\appendix

\end{document}